\documentclass[journal,twoside,web]{ieeecolor}
\usepackage{tmi}
\usepackage{cite}
\usepackage{amsmath,amssymb,amsfonts}
\usepackage{algorithmic}
\usepackage{graphicx}
\usepackage{textcomp}
\usepackage{multirow}
\usepackage{stfloats}

\definecolor{darkcyan}{RGB}{0,138,218}
\usepackage[colorlinks=true, allcolors=darkcyan,urlcolor=black]{hyperref}

\usepackage[switch]{lineno}
\usepackage{bbding}

\def\BibTeX{{\rm B\kern-.05em{\sc i\kern-.025em b}\kern-.08em
    T\kern-.1667em\lower.7ex\hbox{E}\kern-.125emX}}
\begin{document}

\title{Asymmetric Co-Training with Explainable Cell Graph Ensembling for Histopathological Image Classification}

\author{Ziqi Yang, Zhongyu Li, Chen Liu, Xiangde Luo, Xingguang Wang, Dou Xu, Chaoqun Li, Xiaoying Qin, Meng Yang, Long Jin
\thanks{This study was supported by grants from the High-level Hospital Foster Grants from Fujian Provincial Hospital of China (grant number 2019HSJJ25) and the Fujian Provincial Natural Science Foundation Project of China (grant number 2022J01406).}
\thanks{Z. Yang, Z. Li, and C. Liu contributed equally. Corresponding author: L. Jin (Email: Jinlongdoctor@163.com).}
\thanks{Z. Yang, Z. Li, X. Wang, D. Xu, and C. Li are with the School of Software Engineering, Xi'an Jiaotong University, Xi'an, China.}
\thanks{C. Liu, X. Qin, and L. Jin are with the Department of Pathology, Shengli Clinical Medical College, Fujian Medical University, Fuzhou, Fujian, China.}
\thanks{X. Luo is with the School of Mechanical and Electrical Engineering, University of Electronic Science and Technology of China, Chengdu, China.}
\thanks{M. Yang is with the Hunan Frontline Medical Technology Co., Ltd.}}

\maketitle

\begin{abstract}
Convolutional neural networks excel in histopathological image classification, yet their pixel-level focus hampers explainability. Conversely, emerging graph convolutional networks spotlight cell-level features and medical implications. However, limited by their shallowness and suboptimal use of high-dimensional pixel data, GCNs underperform in multi-class histopathological image classification. To make full use of pixel-level and cell-level features dynamically, we propose an asymmetric co-training framework combining a deep graph convolutional network and a convolutional neural network for multi-class histopathological image classification. To improve the explainability of the entire framework by embedding morphological and topological distribution of cells, we build a 14-layer deep graph convolutional network to handle cell graph data. For the further utilization and dynamic interactions between pixel-level and cell-level information, we also design a co-training strategy to integrate the two asymmetric branches. Notably, we collect a private clinically acquired dataset termed LUAD7C, including seven subtypes of lung adenocarcinoma, which is rare and more challenging. We evaluated our approach on the private LUAD7C and public colorectal cancer datasets, showcasing its superior performance, explainability, and generalizability in multi-class histopathological image classification.
\end{abstract}


\section{Introduction}
\label{sec:introduction}
\IEEEPARstart{H}{istopathological} image classification plays an important role in medical-decision-makings. According to the Global Cancer Statistics 2018, lung cancer is the leading cause of cancer morbidity and mortality \cite{bray2018global}, accounting for approximately one-fifth of cancer deaths, with lung adenocarcinoma (LUAD) being the most common histological type. Colorectal cancer (CRC) is the third most common cancer and the second most common cause of cancer-related mortality \cite{bray2018global}. Different classifications are significantly related to the prognosis of patients. Recently, deep learning methods have already achieved impressive performance, ranging from early screening to determining subcategories \cite{liu2021co}.

Deep learning approaches \cite{kanavati2020weakly,xue2022robust,xu2019attention,qaiser2019learning,setio2017validation,litjens2017survey} based on convolutional neural networks (CNN) have been widely applied to histopathological image analysis due to their significant performance on various medical imaging tasks. However, such advantages come at the cost of reduced transparency in decision-making processes \cite{tizhoosh2018artificial,hagele2020resolving}. Doctors make decisions by observing the morphology and tissue structure of cells in the histopathological image when diagnosing cancer. But CNN-based methods process histology images at the pixel level, ignoring the notion of histologically meaningful entities, i.e., the specific morphology and tissue structure information at the cell level. The inattention to established cell-level prior pathological knowledge severely limits the explainability of the CNN-based diagnostic frameworks.

In recent years, the surge of graph convolutional networks (GCN) \cite{zhou2019cgc,pati2022hierarchical}, a branch of deep learning characterized by graph-level model development, has brought a new wave of information fusion techniques through their widespread applications in medical image analysis. GCNs have demonstrated their capability to perform feature aggregation, interaction, and reasoning with remarkable flexibility and efficiency \cite{ding2022graph}. Compared to CNN-based methods, GCN-based methods have much better medical explainability. To emphasize the possible medical significance derived from histopathology, cell-based GCNs extract features from the cell level, including the specific cell morphology and tissue structure information. To avoid the problem of feature extraction ability degradation due to sampling diversity, such as chromatic aberration, GCNs handle cell-based graph data relying on topology information rather than pixel information. However, due to the smaller inter-class distance of cell-level features among specific subdivided cancer subtypes, e.g., lung adenocarcinoma, GCN-based methods suffer from poor end-to-end classification performance in multi-class histopathological image classification. Furthermore, GCNs are usually limited to very shallow models due to the vanishing gradient problem, making it difficult to use the high-dimensional pixel information extracted deeply like CNNs. In summary, CNN-based methods lack good explainability, while GCN-based methods suffer from poor performance in multi-class histopathological image classification. The pixel-level and cell-level information of histopathological images has not been comprehensively utilized in an effective way by existing methods.

This paper addresses the aforementioned limitations by transitioning from isolated pixel-based or cell-based approaches to a unified strategy that dynamically harnesses both pixel-level and cell-level features. We propose an asymmetric co-training framework combining an explainable Deep GCN and a high-performance CNN for histopathological image classification. Specifically, to extract high-dimensional information from pixel-level features, the CNN flow regards the total patch image as the input. To extract cell-level features for medical significance, a 14-layer Deep GCN is built to process the cell graph data constructed from image-mask pairs, where the cell nuclei are regarded as the nodes and the potential cellular interactions as the edges. To get masks from the images without extra cell annotations, we train a  cell segmentation network with unsupervised domain adaptation. Finally, to integrate the two asymmetric branches, i.e., the CNN flow and the Deep GCN flow dynamically and make them learn from each other, we use Kullback-Leibler (KL) divergence with a  threshold plus two different losses to constrain the two models for co-training. Within our framework, the CNN primarily addresses multi-class histopathological image classification due to its superior performance, while the Deep GCN augments explainability by embedding morphological and topological distribution of cells. Consequently, our framework synergistically exploits and interacts with both pixel-level and cell-level information, enhancing classification performance and medical explainability concurrently.
The main contributions of this paper can be summarized as follows:
\begin{itemize}
\item We propose an asymmetric co-training framework combining Deep GCN and CNN for the comprehensive utilization and dynamic interactions of pixel-level and cell-level features to improve both model performance and explainability in multi-class histopathological image classification.
\item We build a 14-layer Deep GCN to handle cell-based graph data, for the first time regarded as a complementary model to make the entire framework more explainable by embedding morphological and topological distribution of cells.
\item We collect a private clinically acquired dataset called LUAD7C, including seven subtypes of lung adenocarcinoma, which is rare and challenging in multi-class histopathological image classification.
\item The proposed method has been validated on two typical multi-class histopathological image classification datasets, i.e., our private LUAD7C dataset and the public CRC dataset. The experiments show that our proposed framework consistently outperforms other state-of-the-art CNN-based and GCN-based methods, demonstrating its effectiveness and generalization.
\end{itemize}

\section{Related Work}
Following a divide-and-conquer strategy, this paper explicitly decouples the multi-class histopathological image classification task into two asymmetric sub-tasks, including the CNN pipeline and the Deep GCN pipeline. Notably, to get masks from the images without extra cell annotations, we train an unsupervised domain adaptation cell segmentation network. In the following parts, we will introduce the related works of cell segmentation in histopathological images and multi-class histopathological image classification including CNN-based methods and GCN-based methods.

\subsection{Cell Segmentation In Histopathological Images}
Recent years have witnessed significant exploration of cell segmentation within histopathological images using diverse deep learning methods. Ronneberger et al.\cite{ronneberger2015u} proposed U-Net, which employs a multi-layer deconvolution network to aggregate results, significantly enhancing learning and prediction capabilities. Chen et al.\cite{chen2016dcan} proposed DCAN, utilizing a multi-task learning framework for improved segmentation, encompassing probability maps and clear contours.
However, the scarcity of well-annotated datasets for cell segmentation in clinical diagnosis is notable. To address this, domain-adaptive segmentation techniques have been proposed, mitigating manual annotation efforts. Haq and Huang\cite{haq2020adversarial} proposed a GAN-based framework, coupled with a reconstruction network, for segmenting unlabeled data from diverse organs. Li et al.\cite{li2021unsupervised} proposed a novel framework incorporating semi-supervised segmentation with domain adaptation in the target domain.
In this paper, we focus on training an unsupervised domain adaptation cell segmentation network to derive masks from histopathological image patches without requiring additional cell annotations. Thus, the cell graph and cell adjacency matrix are further constructed as the input of Deep GCN.

\subsection{CNN-Based Medical Image Classification}
The rapid development of deep learning theory has provided a technical approach to solving medical image classification tasks. Most methods use convolutional neural networks to analyze pathological images at the pixel level. Han et al.\cite{han2017breast} leverage hierarchical feature representation for breast cancer multiclassification. Coudray et al.\cite{coudray2018classification} trained a deep learning model to classify and predict mutations from non–small cell lung cancer histopathology. Tellez et al.\cite{tellez2019neural} proposed CNN methods to address trade-offs by leveraging visual context. Though CNN-based methods have achieved notable performance in histopathological image classification tasks, they do not have good explainability due to their black box properties. CNN-based methods operate on fix-sized input patches, confine their field of view, and hinder the incorporation of information from various spatial distances.  Further, pixel-based processing in CNNs disregards the notion of histologically meaningful entities like cells.

\subsection{GCN-Based Medical Image Classification}
In recent years, GCN-based methods have gained prominence among researchers for their capability to facilitate the analysis of cellular characteristics within histopathological images, encompassing nuclear appearance and cell-level spatial information. Kipf and Welling\cite{kipf2016semi} proposed a graph convolutional network that uses an efficient layer-wise propagation rule that is based on a first-order approximation of spectral convolutions on graphs. CGC-Net proposed by Zhou et al.\cite{zhou2019cgc} is a pioneering GCN for cancer grading that bridges the gap between the deep learning framework and the conventional cell graph. It utilizes the nuclei rather than small patches as descriptors. Li et al.\cite{li2019deepgcns} proposed DeepGCNs to solve the problem of gradient disappearance caused by too many GCN layers. Extending beyond cell-level graphs, Pati et al.\cite{pati2022hierarchical} proposed a HACT for histopathological image analysis, incorporating a tissue-level graph established through the super-pixel technique. As opposed to CNN-based methods, GCN-based methods emphasize the possible biological significance derived from histopathology. Despite good explainability on clinical classification tasks, these approaches do not work well in capturing the diagnostic and prognostic information from the surrounding micro-environments at the pixel level, resulting in low performance in multi-class histopathological image classification.

\section{Method}

\subsection{Overview}
\begin{figure}[tp]
\centerline{\includegraphics[width=\columnwidth]{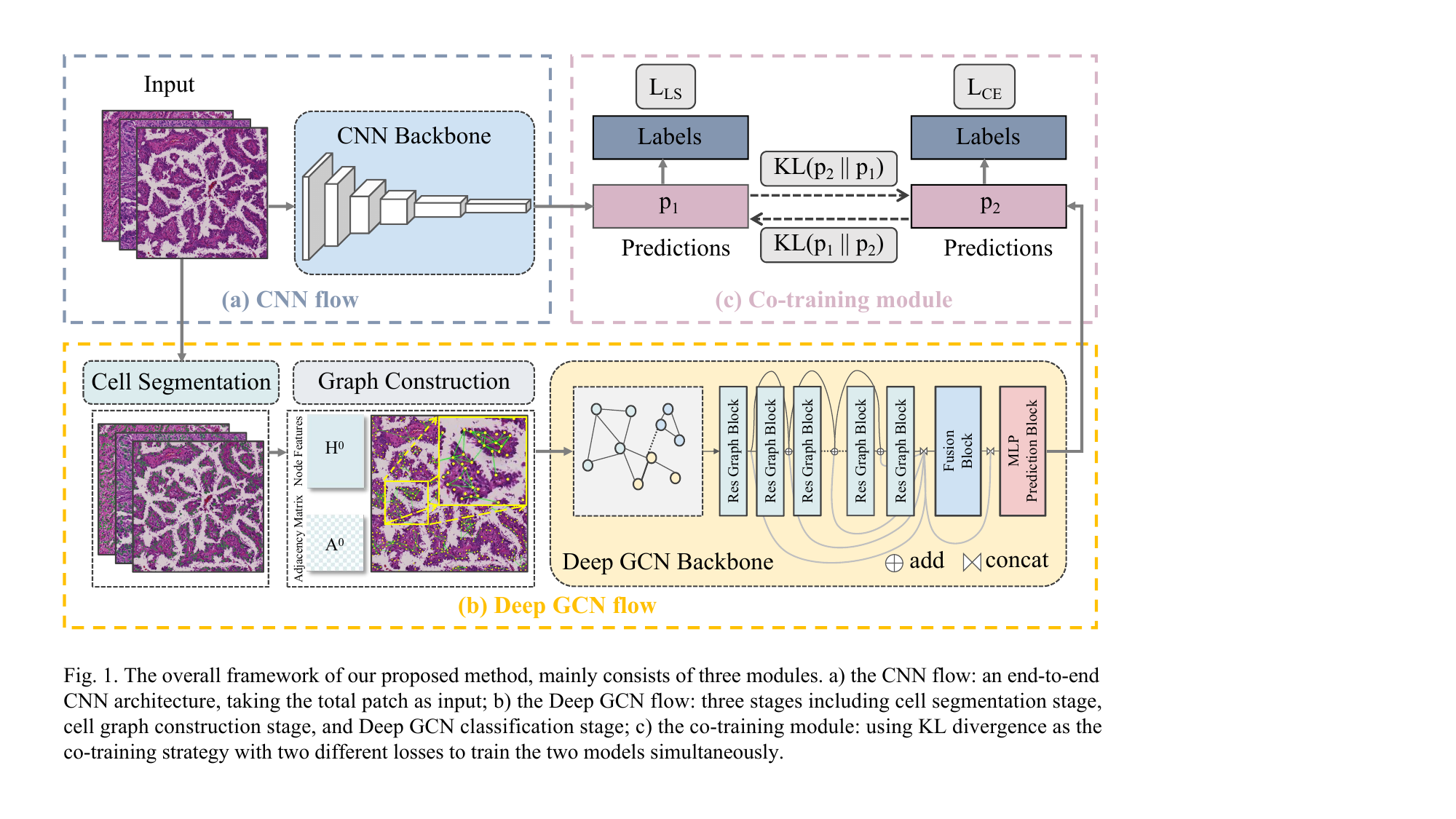}}
\caption{The overall framework of our proposed method, mainly consists of three modules. a) the CNN flow: an end-to-end CNN architecture, taking the total patch as input; b) the Deep GCN flow: three stages including cell segmentation, cell graph construction, and Deep GCN classification; c) the co-training module: using the KL divergence with a constraint threshold and two different losses to train the two models simultaneously.}
\label{fig1}
\end{figure}

As shown in \autoref{fig1}, we present an asymmetric co-training framework for the explainable classification of histopathological images, which combines pixel-level features from convolutional neural networks and cell-level features from deep graph convolutional neural networks, i.e., the CNN flow and the GCN flow. To extract pixel-level features, the CNN flow takes the total patch as input and then gives its predictions using a CNN backbone. To extract cell-level features, the Deep GCN flow is divided into three stages including the cell segmentation, the cell graph construction, and the Deep GCN classification. Finally, to integrate the pixel-level and the cell-level features from the above two asymmetric flows, a co-training module is proposed to constrain two different models in learning complementary information, as well as keeping their independent learning ability. Especially, this asymmetric co-training is embedded using the Kullback-Leibler divergence with a constraint threshold and two different loss functions.

In the following parts, we will give a detailed explanation of the cell segmentation network architecture, cell graph construction process, Deep GCN architecture, co-training module, and the loss functions we used during training. Our motivation will also be explained.

\subsection{Domain Adaptative Cell Segmentation}
\begin{figure}
\centerline{\includegraphics[width=\columnwidth]{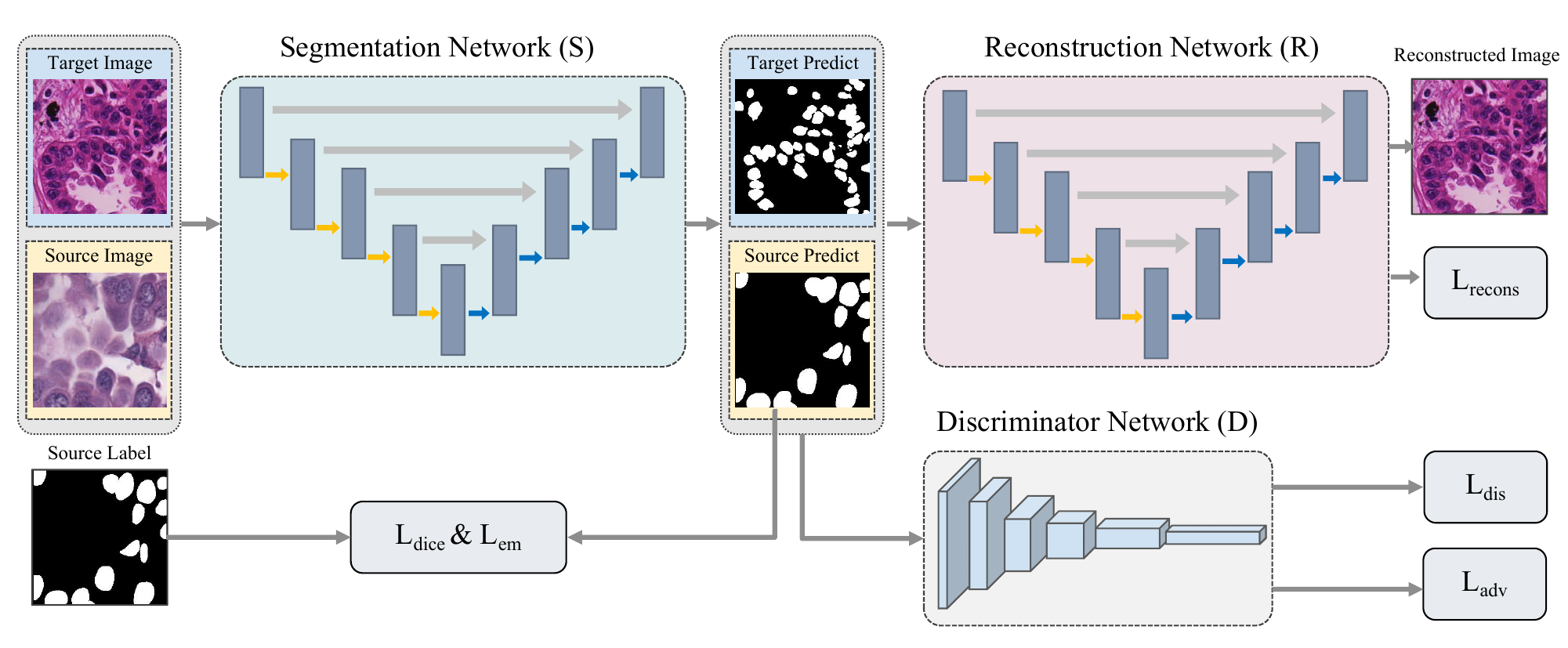}}
\caption{The unsupervised domain adaptative cell segmentation architecture consists of three modules. Segmentation Network (S) generates segmentation masks, from which Reconstruction Network (R) reconstructs input images. Discriminator (D) distinguishes between source domain outputs and target domain outputs from Segmentation Network (S).}
\label{fig2}
\end{figure}

Accurate cell instance segmentation leads to more reliable node features in the cell graph \cite{zhou2019cgc}. To train a cell segmentation network in a fully supervised manner for a particular cancer, we need to label all cells. However, this task is substantial and impractical due to its immense workload. To get masks from the target images without extra cell annotations, we designed an unsupervised domain adaptation cell segmentation network. \autoref{fig2} illustrates its architecture, which can be divided into three modules: Segmentation Network (S), Reconstruction Network (R), and Discriminator (D). The training process necessitates the inclusion of our unlabeled target dataset and a labeled source public dataset (e.g., TCIA dataset \cite{hou2020dataset}).

\textbf{Segmentation network (S):} Our segmentation network $\text S$ takes image patches as input $X$ of size $H \times W \times C$, and produces the segmentation prediction $\hat{Y}$ of the same size as input, i.e., $\hat{Y} = S(X)$. We train $\text S$ to generate the segmentation predictions $\hat{Y}_s$ similar to the ground-truth labels ${Y}_s$ in the source domain. We use both the dice-coefficient loss and the entropy minimization loss as our segmentation loss:
\begin{equation}
L_{dice}=1-\frac{2 \cdot Y_s^{\prime} \cdot \hat{Y}_{\mathrm{s}}^{\prime}}{Y_s^{\prime}+\hat{Y}_{\mathrm{s}}^{\prime}}
\end{equation}

\begin{equation}
L_{e m}=-\frac{1}{H \times W} \sum_{h=0}^H \sum_{w=0}^W \hat{Y}_s \log (\hat{Y}_s)
\end{equation}
where $Y_s^{\prime}$ and $\hat{Y}_{\mathrm{s}}^{\prime}$ are flatten ${Y}_s$ and $\hat{Y}_s$ respectively. To make the distribution of target predictions $\hat{Y}_t$ closer to source predictions $\hat{Y}_s$. Thus, we define adversarial loss as:
\begin{equation}
L_{a d v}\left(X_t\right)=-\frac{1}{H^{\prime} \times W^{\prime}} \sum_{h^{\prime}, w^{\prime}} \log (D(\hat{Y}_t))
\end{equation}
where $\hat{Y}_t = S(X_t)$, and $H^{\prime}$ and $W^{\prime}$ are height and width of discriminator output $D(\hat{Y}_t)$ respectively. This segmentation network can be considered as the generator module of a GAN \cite{goodfellow2020generative}. And the adversarial loss helps $\text S$ to fool the discriminator so that it considers $\hat{Y}_t$ as source domain segmentation masks.

\textbf{Reconstruction Network (R):} We consider $ \text S$ as an encoder and $\text R$ as a decoder to reconstruct original images from predictions. It takes the predictions $\hat{Y}_t$ as inputs and produces the reconstructed image as the output $R(\hat{Y}_t)$. We calculate the reconstruction loss as:
\begin{equation}
L_{r e c o n s}\left(X_t\right)=\frac{1}{H \times W \times C} \sum_{h, w, c}(X_t-R(\hat{Y}_t))^2
\end{equation}
This reconstruction network is used to ensure the correspondence between the target domain predictions and the target input images.

\textbf{Discriminator (D):} To generate similar predictions for both source images and target images, a Discriminator is built. It can take the predictions as its input and distinguish whether the input comes from the source domain or the target domain. To train $\text D$, we use following cross-entropy loss:
\begin{equation}
\begin{aligned}
L_{dis}(\hat{Y})=
&-\frac{1}{H^{\prime} \times W^{\prime}} \sum_{h^{\prime}, w^{\prime}} z \cdot \log (D(\hat{Y})) \\
&+(1-z) \cdot \log (1-D(\hat{Y}))
\end{aligned}
\end{equation}
where $z=0$ when $\text D$ takes target domain prediction as its input, and $z=1$ when input comes from source domain prediction. Overall, we optimize the following total loss when training our framework:
\begin{equation}
\begin{aligned}
L\left(X_s, X_t\right)=
&L_{dice}\left(X_s\right)+L_{em}\left(X_s\right)+\lambda_{adv}L_{adv}\left(X_t\right)\\
&+\lambda_{recons} L_{recons}\left(X_t\right)+L_{dis}(\hat{Y})
\end{aligned}
\end{equation}
where the $\lambda_{adv}$ and $\lambda_{ recons}$ are the weights to balance above losses.

\subsection{Cell Graph Construction}
It's of vital importance to construct a meaningful cell graph including efficient node features and a representative adjacency matrix from the image-mask pairs obtained by the above domain adaptative cell segmentation. To utilize both cell appearance and spatial information, we develop a cell graph construction pipeline that integrates cell-level feature and position information. More specifically, the cell graph construction stage can be divided into the following steps: i) cell feature extraction to define node features; ii) representative cell sampling to remove redundancy in the graph; iii) graph edge configuration to define potential cellular interactions.

\textbf{Cell Feature Extraction:} In a cell graph, the segmented cell nuclei are considered nodes. The node attribute is defined as the combination of image wised features, here means the hand-crafted features. Referring to previous practices by \cite{zhou2019cgc}, we choose the 16 most predictive cell-level features: mean nuclei intensity; orientation; solidity; cell perimeter; minimum length of axis; maximum length of axis; area; eccentricity; average fore-/background difference; the standard deviation of nuclei intensity; skewness of nuclei intensity; mean entropy of nuclei intensity; GLCM of dissimilarity; GLCM of homogeneity; GLCM of ASM;  GLCM of energy. In addition, we incorporate the centroid coordinates. Therefore we in total adopt sixteen hand-crafted features and one cell centroid coordinate as our node descriptors, which ensemble cell-level feature and position information.

\textbf{Representative Cell Sampling:} In the step of the cell selection, utilizing all cells in the image as nodes is unnecessary because some regions are dense with many cells containing similar features. To resolve this problem, we used a fusion cell sampling strategy, thus making the selected cells more representative, instead of using all of them. To be specific, we fuse the Farthest Point Sampling (FPS) method \cite{eldar1997farthest} and the random sampling method.

\textbf{Graph edge configuration:} The previous method of constructing a cell adjacency matrix is to calculate the Euclidean distance between cells. However, considering that tumor cells and normal ones have distinct feature differences, this paper integrates the feature and position information as the overall features to calculate the distance in feature space. The overall features can be defined as follows:
\begin{equation}
t=[\alpha \times g, \beta \times c]
\end{equation}
where $t$ represents the overall features, $\alpha$ and $\beta$ are weight parameters, respectively, $g$ is the extracted 16-dimensional feature, and $c$ is the coordinate information of the cell. The specific method of constructing the adjacency matrix is as follows:
\begin{equation}
\begin{aligned}
A_{i j}=\left\{\begin{array}{cc}
1 & \text { if } j \in K N N(i) \text { and } D({t}_i, {t}_j)<d, \\
0 & \text {otherwise.}
\end{array}\right.
\end{aligned}
\end{equation}
where $A$ represents the cell adjacency matrix, $KNN(i)$ is the k-nearest neighbors of node $i$, and $D({t}_i, {t}_j)$ denotes the overall features distance between node $i$ and node $j$, and $d$ is the distance threshold.

\begin{figure}
\centerline{\includegraphics[width=\columnwidth]{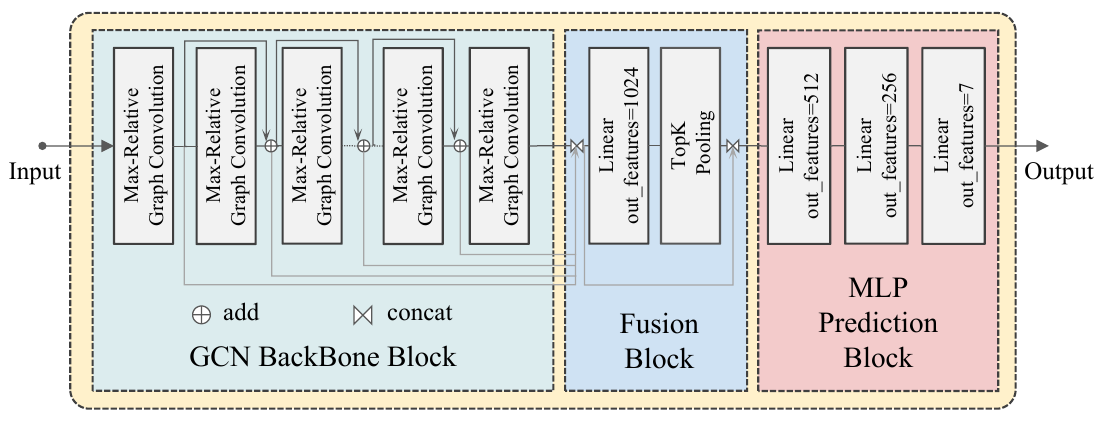}}
\caption{The deep graph convolutional network architecture: a GCN backbone block (ResGCN architecture is shown here), a fusion block, and an MLP prediction block.}
\label{fig3}
\end{figure}

\subsection{Deep Graph Convolutional Network Architecture}
After the cell graph construction stage, we aim to extract cell-level features deeply from the cell graphs. To improve the explainability of the entire framework by embedding morphological and topological distribution of cells, we build a 14-layer Deep GCN to process the cell graph data constructed from image-mask pairs. To solve the problem that most previous arts employ shallow GCNs that are usually no more than 4 layers deep \cite{zhou2020graph} caused by high complexity in back-propagation of stacking multiple layers of graph convolutions \cite{li2018deeper,zhou2020graph,wu2020comprehensive}, we train a 14-layer deep GCN by adapting residual/dense connections and dilated convolutions \cite{yu2015multi}, which is first applied in multi-class histopathological image classification.

Unlike traditional convolution in the CNN flow which operates on the regular grid in the Euclidean space, the graph convolution extends the information aggregation to the non-Euclidean space, perfectly adapted to the irregular cell graph structure. Different from the conventional graph learning framework that learns a direct underlying mapping, which takes a graph as an input and outputs a new graph representation, the Deep GCN we build consists of graph residual learning blocks. Specifically, we insert shortcut connections after the normal mapping.

As illustrated in \autoref{fig3}, the deep graph convolutional network architecture consists of three blocks: a GCN backbone block, a fusion block, and an MLP prediction block. Here we choose ResGCN as the GCN backbone. The input cell graph is represented by a node feature matrix and an adjacency matrix. ResGCN takes full use of residual skip connections for all GCN layers. The node features extracted from the GCN backbone blocks at each GCN layer are concatenated and taken as the input of the fusion block, which is used to fuse the global and multi-scale local features. Inside the fusion block is a linear layer followed by a TopK pooling. The previous layer maps the features to a fixed-dimensional feature space. The latter layer aggregates the node features of the entire graph into a single global feature vector which represents global information. Then we concatenate the global features with the extracted features of each node from all previous GCN layers in order to fuse global and local information. Finally, the MLP prediction block applies three MLP layers to make predictions using the fused comprehensive features.

\subsection{Asymmetric Co-Training Strategy}
To solve the feature fusion noise problem caused by two different data modalities (e.g., image data and cell-graph data) of conventional model fusion methods like add, multiplication, and concatenation forms, we propose an asymmetric co-training strategy for the comprehensive utilization of the feature extraction capabilities of the two branch models. Specifically, we use KL divergence with a threshold and two different losses to train two models simultaneously. In this way, the pixel-level feature information extracted by the CNN model and the cell-level feature information extracted by the Deep GCN model can be fused in a dynamic way instead of being concatenated in a static way. This strategy can enhance the interaction between the two asymmetric branch models, which leads to improving the performance of both models \cite{blum1998combining,zhang2018deep}.

The proposed asymmetric co-training architecture with a cohort of two type networks (see \autoref{fig1}) can be formulated as follows. Given $N$ samples $\mathcal{X}=\left\{\boldsymbol{x}_i\right\}_{i=1}^N$ from $M$ classes, we denote the corresponding label set as $\mathcal{Y}=\left\{y_i\right\}_{i=1}^N$ with $y_i \in\{1,2, \ldots, M\}$. The CNN model and the Deep GCN model are named $\Theta_1$ and $\Theta_2$,  respectively. The probability of class $m$ for sample $x_i$ given by a neural network $\Theta_1$ is computed as: 
\begin{equation}
p_1^m\left(\boldsymbol{x}_i\right)=\frac{\exp \left(z_1^m\right)}{\sum_{m=1}^M \exp \left(z_1^m\right)}
\end{equation}
where the logit $z^m$ is the output of the “softmax” layer in $\Theta_1$. In order to improve the generalization and calibration of the CNN model, we take the label smoothing method \cite{szegedy2016rethinking} as its loss function, which encourages the representations of training examples from the same class to group in tight clusters \cite{muller2019does}. It can be formulated as follows:
\begin{equation}
L_{LS}=-\sum_{i=1}^N \sum_{m=1}^M I\left(y_i, m\right) \log \left(p_1^m\left(\boldsymbol{x}_i\right)\right)
\end{equation}
with an indicator function $I$ defined as:
\begin{equation}
I\left(y_i, m\right)=\left\{\begin{array}{l}
1-\alpha, \text { if } y_i=m \\
\frac{\alpha}{M-1}, \text { if } y_i \neq m
\end{array}\right.
\end{equation}
where $\alpha$ Indicates the label smoothing factor. To further improve the generalization performance of $\Theta_1$ on the testing instances, we use another asymmetric network $\Theta_2$ to provide training experience in the form of its posterior probability $p_2$. The network $\Theta_2$ here means the explainable Deep GCN model, whose loss function is a conventional cross-entropy loss $L_{CE}$. To quantify the match of the two network’s predictions $p_1$ and $p_2$, we use the Kullback-Leibler divergence. The KL distance from $p_1$ to $p_2$ is computed as:
\begin{equation}
D_{K L}\left(\boldsymbol{p}_2 \| \boldsymbol{p}_1\right)=\sum_{i=1}^N \sum_{m=1}^M p_2^m\left(\boldsymbol{x}_i\right) \log \frac{p_2^m\left(\boldsymbol{x}_i\right)}{p_1^m\left(\boldsymbol{x}_i\right)}
\end{equation}
It is worth noting that in order to enable the models of two different modalities to learn complementary information while maintaining a certain independent learning ability, we set a threshold $d_{KL}$ to constrain the KL divergence. The loss functions $L_{\Theta_1}$ and $L_{\Theta_2}$ for networks $\Theta_1$ and $\Theta_2$ respectively are thus:
\begin{equation}
\begin{aligned}
&L_{\Theta_1}=L_{LS}+\lambda(\boldsymbol{p}_2 \| \boldsymbol{p}_1)D_{K L}\left(\boldsymbol{p}_2 \| \boldsymbol{p}_1\right) \\
&L_{\Theta_2}=L_{CE}+\lambda(\boldsymbol{p}_1 \| \boldsymbol{p}_2)D_{K L}\left(\boldsymbol{p}_1 \| \boldsymbol{p}_2\right)
\end{aligned}
\end{equation}
where $\lambda(\boldsymbol{p}_i \| \boldsymbol{p}_j)$ is defined as:
\begin{equation}
\lambda(\boldsymbol{p}_i \| \boldsymbol{p}_j)=\left\{\begin{array}{l}1, 
\text { if } D_{K L}(\boldsymbol{p}_i \| \boldsymbol{p}_j) \geq d_{KL} \\0, 
\text { if } D_{K L}(\boldsymbol{p}_i \| \boldsymbol{p}_j) < d_{KL}\end{array}\right.
\end{equation}
The overall asymmetric co-training loss function $L_{ACT}$ is thus: 
\begin{equation}
\begin{aligned}
L_{ACT} &= L_{\Theta_1} + L_{\Theta_2} \\
\end{aligned}
\end{equation}

The co-training module composed of all the losses above and the training strategy together greatly contributes to the dynamic interactions between the two asymmetric flows. Also, it is worth mentioning that extension to more networks is possible following our asymmetric co-training architecture and strategy.

\section{Experiments}

\subsection{Datasets}

\subsubsection{LUAD7C Dataset}
We first evaluated our method on our private clinically acquired LUAD7C dataset\footnote{Our LUAD7C dataset will be made publicly available subsequently.}, encompassing seven subtypes of lung adenocarcinoma. The dataset contains 94 Whole Slide Images (WSI) stained with H\&E, including seven subclasses of lung adenocarcinoma (LUAD). All WSIs were extracted from 94 different patients collected at Fujian Provincial Hospital and were scanned employing the professional HS6 pathological slide scanner at $40\times$ magnification. It was collected and annotated clinically by experienced professional pathologists, which took more than a year. The ROIs of the seven subclasses of LUAD were annotated by two pathologists with over five years of experience. All the annotations were checked by a third pathologist with over ten years of experience. Difficult and disagreement samples were decided by three experts together. Specifically, the dataset contains seven histopathological types of lung adenocarcinoma tumors: Acinar adenocarcinoma (AAC), Lepidic adenocarcinoma (LAC), Papillary adenocarcinoma (PAC), Micropapillary adenocarcinoma (MPAC), Solid adenocarcinoma (SAC), Invasive mucinous adenocarcinoma (IMAC), and Complex glandular pattern (including cribriform pattern) (CGP).
The vast dimensions of LUAD7C WSIs with an average resolution of $80,000\times60,000$ pose computational and algorithmic challenges. Additionally, the presence of diverse tumors from various LUAD subtypes within a single WSI necessitates preprocessing. Raw LUAD7C data undergoes segmentation into uniformly sized patch images with annotations, resulting in 13,927 non-overlapping patches, each measuring $256\times256$ pixels. For experimentation purposes, the LUAD7C dataset is randomly divided into training (80\%) and testing (20\%) subsets on a patch basis. These subsets stem from distinct WSI sets and maintain a balanced distribution across seven classes.

\subsubsection{CRC Dataset} 
Our method's generalization capability was further assessed using the publicly available CRC dataset \cite{awan2017glandular}, characterized by three distinct labels associated with gland differentiation levels. This dataset encompasses distinct, non-overlapping images measuring $4,548\times7,548$ pixels, captured at $20\times$ magnification. Expert pathologists have categorized each image into normal tissue, low-grade tumors, or high-grade tumors. The dataset was compiled from digitized WSIs originating from 38 colorectal adenocarcinomas (CRA) tissue slides, subjected to H\&E staining. These WSIs, procured from diverse patients, were scanned with the Omnyx VL120 scanner at 0.275 $\mu m/pixel$. The dataset includes a total of 139 images: 71 normal, 33 low-grade, and 35 high-grade cancer images.
Analogous to the data preprocessing applied to our LUAD7C dataset, we generated 4,734 non-overlapping patches, each sized $512\times512$ pixels, from the CRC dataset. In our experimentation, the CRC dataset was randomly partitioned into training (80\%) and testing (20\%) subsets on a patch basis. These subsets, sourced from distinct sets, exhibit a balanced distribution across the three classes. 

\subsection{Implementation}

\subsubsection{Network Details}
As shown in \autoref{fig1}, our proposed framework contains a CNN model, a cell segmentation network, a Deep GCN model, and a co-training module. For the CNN model, we choose EfficientNet-B0 as our backbone. For the cell segmentation network, U-Net \cite{ronneberger2015u} is adopted as both of Segmentation Network (S) and Reconstruction Network (R), and Discriminator (D) is implemented by five convolutional layers. We use 0.001 and 0.01 as $\lambda_{adv}$ and $\lambda_{recons}$ respectively. For the Deep GCN model, we build a 14-layer ResGCN using PyTorch \cite{paszke2017automatic} with the geometric deep learning package \cite{fey2019fast}. As for the co-training module, we use KL divergence as the co-training strategy with two different losses, including the cross-entropy loss and the label smoothing cross-entropy loss.

\subsubsection{Training Details}
Our proposed framework was trained by minimizing the cross-entropy loss, the label smoothing cross-entropy loss, and the KL divergence loss together. To prevent over-fitting, Dropout \cite{srivastava2014dropout} is used with $p = 0.2$ during training. We use AdamW optimization with an initial learning rate of $2e^{-3}$. The learning rate is reduced to 0.5 times the original, with no improvement after every three consecutive epochs. And the minimum learning rate is set to $1e^{-8}$. The CNN model and Deep GCN model are trained simultaneously for 200 epochs with a batch size of 32. As for the co-training strategy, we set the KL divergence threshold to 0.1 and label smoothing \cite{szegedy2016rethinking} to 0.1. All experiments are carried out without data augmentation using PyTorch \cite{paszke2017automatic} on a Linux system with four RTX 2080Ti GPUs.

\subsection{Evaluation Metrics}
We use two metrics to evaluate the performance of models: Accuracy and F1-score. Accuracy evaluates the performance of the entire model, while the F1-score evaluates both the Precision and Recall comprehensively. For multi-classification tasks, we compute the Precision, Recall, F1-score for each class.

\subsection{Comparison with State-of-the-Art Methods}
To prove the performance of the proposed framework, we first compare the proposed method with state-of-the-art methods, including 4 CNN-based methods, i.e., ResNet\cite{he2016deep}, DenseNet\cite{huang2017densely}, MobileNet-V2\cite{sandler2018mobilenetv2}, EfficientNet\cite{tan2019efficientnet}, and 5 GCN-based methods, i.e.,  GCN\cite{kipf2016semi}, ResGCN\cite{li2019deepgcns}, DenseGCN\cite{li2019deepgcns}, CGC-Net\cite{zhou2019cgc}, HACT-Net\cite{pati2022hierarchical}. We train all compared methods on the same training and testing strategy. We report the results of the F1-score per subtype, the average macro F1-score of all categories, and the entire Accuracy. The results are evaluated on the LUAD7C dataset and the CRC dataset, shown in \autoref{table1} and \autoref{table2}, respectively. The best ones are labeled in bold. It can be observed that our proposed method achieves the best result compared with state-of-the-art methods in multi-class histopathological image classification tasks. Also, the results indicate that our LUAD7C dataset is comparatively much more challenging than the CRC dataset.

\begin{table}[tp]
\centering
\caption{The performance comparison with SOTA methods on the LUAD7C dataset.}
\renewcommand\arraystretch{1.2}{
\resizebox{\columnwidth}{!}{
\begin{tabular}{lllllllllll}
\hline
\multicolumn{2}{l}{\multirow{2}{*}{Methods}} &
  \multicolumn{7}{c}{F1-score} & 
  \multirow{2}{*}{\begin{tabular}[c]{@{}l@{}}Macro \\ F1-score\end{tabular}} &
  \multirow{2}{*}{Accuracy} \\ 
\cline{3-9}
\multicolumn{2}{l}{}             & AAC   & LAC   & PAC   & MPAC  & SAC   & IMAC  & CGP   &       &       \\
\hline
\multicolumn{2}{l}{ResNet-18 \cite{he2016deep}}    & 77.62 & 82.67 & 74.97 & 84.32 & 89.09 & 96.40 & 93.59 & 85.52 & 85.13 \\
\multicolumn{2}{l}{DenseNet-121 \cite{huang2017densely}} & 81.26 & 81.65 & 77.66 & 86.39 & 93.37 & 94.58 & 92.60 & 86.79 & 86.50 \\
\multicolumn{2}{l}{MobileNet-V2 \cite{sandler2018mobilenetv2}} & 75.51 & 80.00 & 70.59 & 81.15 & 86.22 & 96.99 & 93.18 & 83.38 & 82.98 \\
\multicolumn{2}{l}{EfficientNet-B0 \cite{tan2019efficientnet}} &
  85.68 &
  88.22 &
  84.23 &
  90.30 &
  \textbf{94.13} &
  97.48 &
  95.65 &
  90.81 &
  90.56 \\ \hline
\multicolumn{2}{l}{GCN \cite{kipf2016semi}}        & 51.53 & 64.86 & 40.89 & 49.84 & 66.94 & 58.18 & 52.07 & 54.90 & 54.33 \\
\multicolumn{2}{l}{ResGCN-14 \cite{li2019deepgcns}}    & 68.76 & 69.84 & 51.60 & 60.52 & 76.96 & 68.98 & 63.90 & 65.79 & 65.21 \\
\multicolumn{2}{l}{DenseGCN-14 \cite{li2019deepgcns}}  & 70.90 & 73.43 & 51.63 & 60.20 & 75.80 & 71.52 & 62.44 & 66.56 & 66.18 \\
\multicolumn{2}{l}{CGC-Net \cite{zhou2019cgc}}      & 56.74 & 69.78 & 48.09 & 55.57 & 68.10 & 65.26 & 52.26 & 59.40 & 58.78 \\
\multicolumn{2}{l}{HACT-Net \cite{pati2022hierarchical}}     & 44.40 & 52.95 & 46.89 & 59.94 & 35.95 & 77.74 & 72.80 & 55.81 & 56.29 \\ \hline
\multicolumn{2}{l}{Ours} &
  \textbf{87.35} &
  \textbf{89.25} &
  \textbf{86.36} &
  \textbf{90.66} &
  93.99 &
  \textbf{98.02} &
  \textbf{96.19} &
  \textbf{91.69} &
  \textbf{91.45} \\
\hline
\end{tabular}}}
\label{table1}
\end{table}

\begin{table}[]
\centering
\caption{The performance comparison with SOTA methods on the CRC dataset.}
\renewcommand\arraystretch{1.1}{
\resizebox{\columnwidth}{!}{
\begin{tabular}{llllll}
\hline
\multirow{2}{*}{Methods} &
  \multicolumn{3}{c}{F1-score} &
  \multirow{2}{*}{\begin{tabular}[c]{@{}l@{}}Macro \\ F1-score\end{tabular}} &
  \multirow{2}{*}{Accuracy} \\ \cline{2-4}
                & Normal & Low-grade & High-grade &       &       \\ \hline
ResNet-18 \cite{he2016deep}       & 97.74  & 97.34     & 97.46      & 97.51 & 97.57 \\
DenseNet-121 \cite{huang2017densely}    & 98.68  & 99.10     & 97.19      & 98.32 & 98.52 \\
MobileNet-V2 \cite{sandler2018mobilenetv2}    & 98.06  & 96.67     & 98.25      & 97.66 & 97.67 \\
EfficientNet-B0 \cite{tan2019efficientnet} & 99.30  & 98.73     & 99.43      & 99.15 & 99.15 \\ \hline
GCN \cite{kipf2016semi}           & 99.40  & 91.46     & 85.71      & 92.19 & 94.61 \\
ResGCN-14 \cite{li2019deepgcns}       & 99.70  & 93.01     & 87.38      & 93.36 & 95.56 \\
DenseGCN-14 \cite{li2019deepgcns}    & 99.60  & 94.57     & 92.49      & 95.55 & 96.83 \\
CGC-Net \cite{zhou2019cgc}         & 98.39  & 91.10     & 85.45      & 91.65 & 93.97 \\
HACT-Net \cite{pati2022hierarchical}        & 98.09  & 97.13     & 97.33      & 97.52 & 97.67 \\ \hline
Ours &
  \textbf{99.90} &
  \textbf{99.82} &
  \textbf{100.00} &
  \textbf{99.91} &
  \textbf{99.89} \\ \hline
\end{tabular}}}
\label{table2}
\end{table}

Typically, GCN-based methods offer superior explainability owing to their biological entity-centric modeling. However, in the context of multi-class histopathological image classification tasks, CNN-based methods generally outperform GCN-based counterparts due to their ability to capture richer and deeper pixel-level features. Our approach transcends this dichotomy by integrating both biological insights in graph structure design and interactions between pixel-level and cell-level entities. As a result, it excels not only over GCN-based approaches but also CNN-based methods on our LUAD7C dataset and the CRC dataset. These results strongly demonstrate the better performance and generalization of our proposed method compared to the existing state-of-the-art methods.

\begin{figure}[tp]
\centerline{\includegraphics[width=\columnwidth]{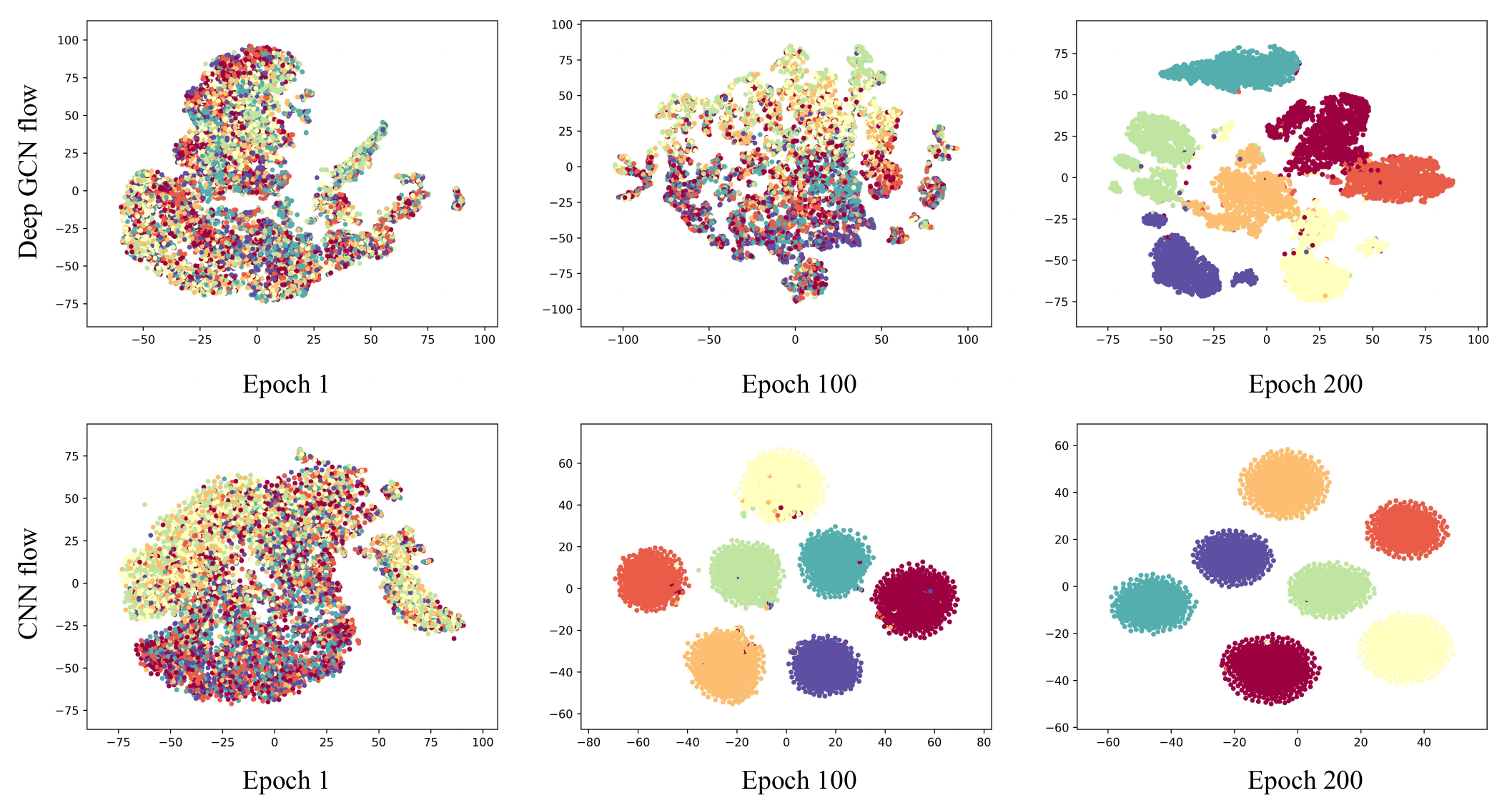}}
\caption{Feature distribution map of different training epochs from 1 to 200 using t-SNE visualization technique, where different colors indicate different subtypes of LUAD7C, the first and second row respectively shows the Deep GCN flow and the CNN flow of our proposed framework during co-training.}
\label{fig4}
\end{figure}

\subsection{Effectiveness of Proposed Framework}
To demonstrate our framework's effectiveness, we monitor the LUAD7C dataset's feature updates across various training epochs (1 to 200). \autoref{fig4} shows the t-SNE plot of feature vectors extracted from different branch models within our framework. The first row shows the feature distribution map of the Deep GCN model (i.e., ResGCN-14) during co-training within the Deep GCN flow. The second row shows the feature distribution map of the CNN model (i.e., EfficientNet-B0) during co-training within the CNN flow. Notably, compared to Deep GCN, the overlapping phenomenon of clusters in the CNN feature space is less, showing a larger inter-class distance and a smaller intra-class distance. Training-wise, CNN outpaces Deep GCN in terms of fitting speed and final classification performance, elucidating CNN's superior efficacy. However, regardless of Deep GCN or CNN, as a component of our unified fused framework, the distinction between the seven cancer subtypes of LUAD7C is obvious and intuitive after training to a certain epoch, further supporting the effectiveness of our proposed framework.

\begin{figure}
\centerline{\includegraphics[width=\columnwidth]{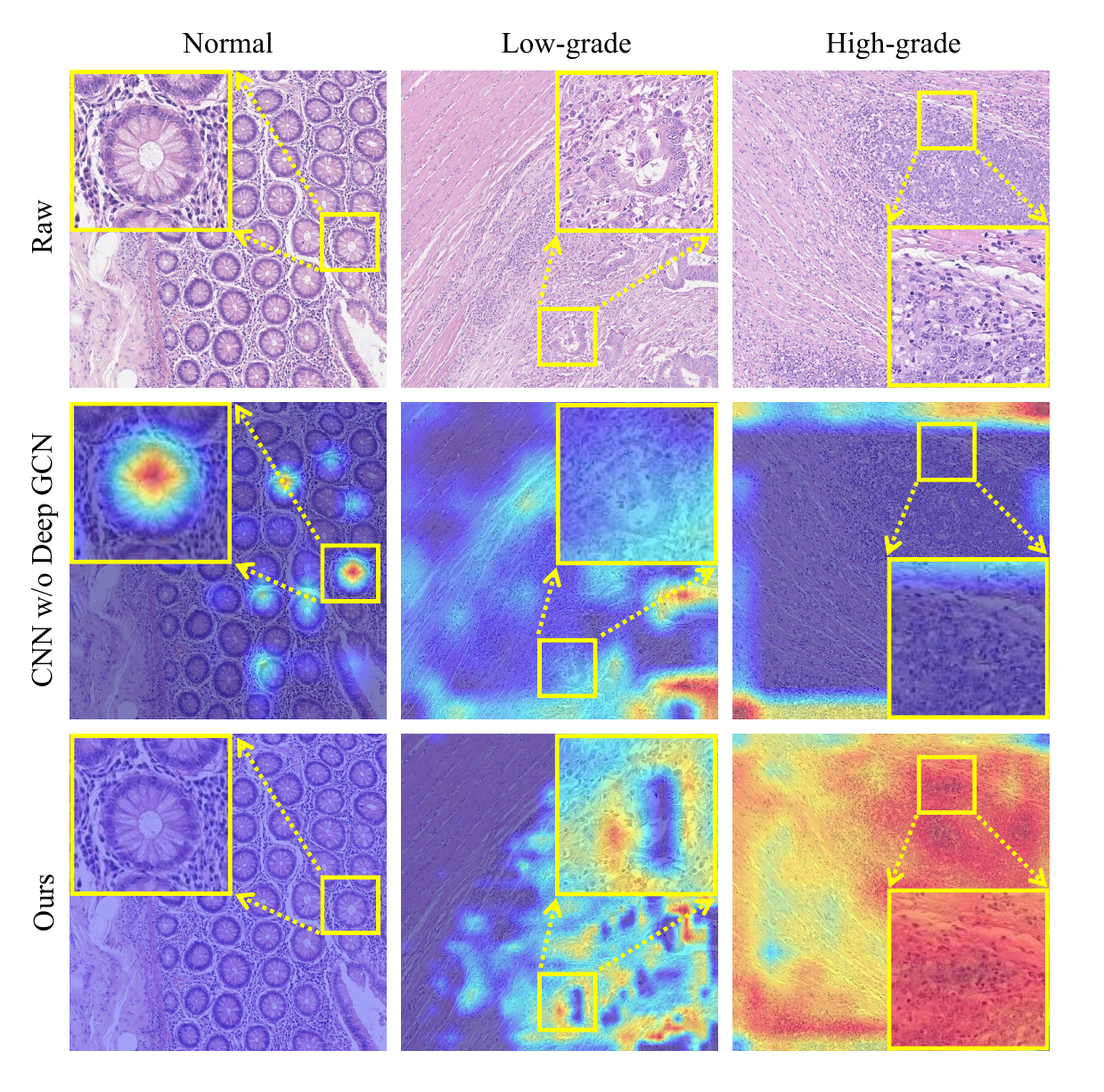}}
\caption{Qualitative explainability on the CRC dataset. The columns represent the three types of CRC tissues, i.e., Normal, Low-grade, and High-grade. The first row shows the raw samples, the second row represents the sample explanations produced by CNN without Deep GCN's co-training, and the third row represents the sample explanations produced by our framework where the auxiliary training of Deep GCN guides CNN to focus on the areas with medical diagnostic significance, making our method more explainable.}
\label{fig5}
\end{figure}

\begin{figure*}[]
\centerline{\includegraphics[width=\textwidth]{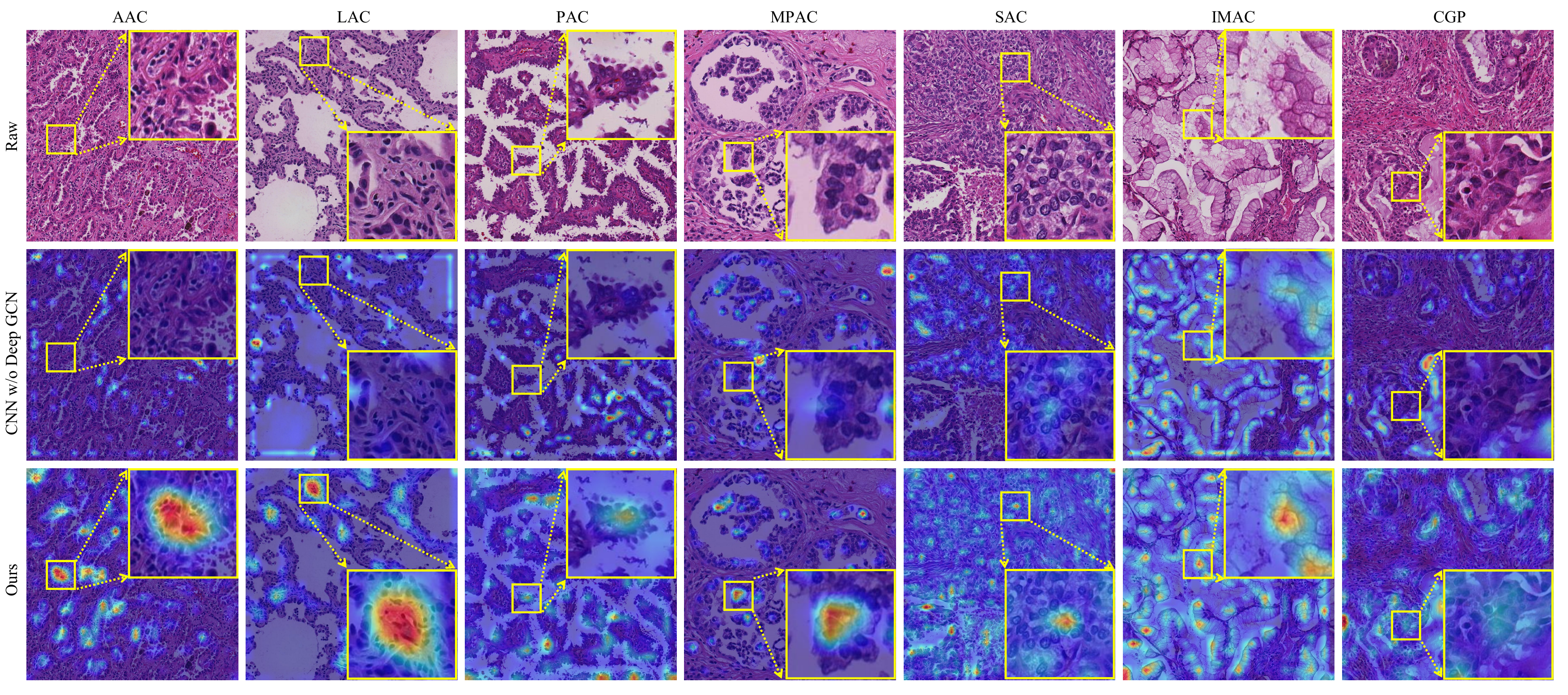}}
\caption{Qualitative explainability on the LUAD7C dataset. The columns represent the seven subtypes of lung adenocarcinoma, i.e., AAC, LAC, PAC, MPAC, SAC, IMAC, and CGP. The first row shows the raw samples of LUAD7C subtypes, the second row represents the sample explanations produced by CNN without Deep GCN's co-training, and the third row represents the sample explanations produced by our framework where the auxiliary training of Deep GCN guides CNN to focus on the areas with medical diagnostic significance, making our method more explainable.}
\label{fig6}
\end{figure*}

\subsection{Explainability of Proposed Framework}
One of the primary obstacles for real-world application of deep learning models in computer-aided diagnosis is the black-box nature of the deep neural networks \cite{adnan2020representation}. In the context of computational pathology, explainability is defined as making the DL decisions understandable to pathologists \cite{holzinger2017towards}.
 
In this work, we adopt a feature attribution approach called GRAD-CAM \cite{selvaraju2017grad}, designed for explaining CNNs operating on images. Specifically, we specify the last layer of the CNN backbone and visualize the importance our framework gives to each pixel for making the final prediction. The deeper the red color, the more significant its contribution to the final classification. Such visualization provides more insight to pathologists regarding the model’s internal decision-making. What's more, in order to show the enhancement effect of Deep GCN on explainability, we respectively visualize and compare the class activation explanation of CNN with and without Deep GCN co-training. The qualitative explainability of the CRC dataset and the LUAD7C dataset are shown in \autoref{fig5} and \autoref{fig6}, respectively.  The magnified yellow box, annotated by pathologists, represents a distinctive pathological area used for characterizing cancer subtypes. The results indicate the pixel-wise visual explanations of CNN without Deep GCN's co-training tend to be blurry and even incorrect. By contrast, the qualitative explanations produced by our framework are much closer to the relevant ROIs and medical landmarks for pathologists to judge cancer subtypes.

Popular deep learning methods disregard biological entities’ notions, thus complicating comprehension by pathologists \cite{jaume2021quantifying}. In our proposed framework, Deep GCN is first transformed as a complementary model to process cell-based graph data emphasizing the notion of biological tissue entities, their topological distribution, and inter-entity interactions. The asymmetric co-training with explainable cell graph ensembling improves the explainability of the CNN model. This enables qualitative post-hoc explanations of the entire framework accessible to pathologists. The explanations allow us to evaluate the pathological relevance of black-box neural network reasoning.

\begin{table}
\centering
\caption{Ablation study on the LUAD7C dataset, tables show the performance improvement of co-training with Deep GCN and CNN compared to training them independently.}
\renewcommand\arraystretch{1.2}{
\resizebox{\columnwidth}{!}{
\begin{tabular}{llllllll}
\hline
\multicolumn{2}{c}{\multirow{2}{*}{Modules}} & \multicolumn{6}{c}{Accuracy / Macro F1-Score of LUAD7C}                                                       \\ \cline{3-8} 
\multicolumn{2}{c}{}                         & \multicolumn{2}{c}{Independent} & \multicolumn{2}{c}{Co-Training} & \multicolumn{2}{c}{Up $\uparrow$} \\ \hline
GCN         & CNN             & GCN          & CNN           & GCN           & CNN           & GCN        & CNN       \\ \hline
ResGCN-14   & ResNet-18       & 65.21 / 65.79  & 85.13 / 85.52   & 65.42 / 66.12   & 85.24 / 85.57   & 0.21 / 0.33  & 0.11 / 0.05 \\
DenseGCN-14 & ResNet-18       & 66.18 / 66.56  & 85.13 / 85.52   & 66.82 / 67.41   & 86.28 / 86.60   & 0.64 / 0.85  & 1.15 / 1.08 \\
ResGCN-14   & EfficientNet-B0 & 65.21 / 65.79  & 90.56 / 90.81   & \textbf{69.55 / 69.95}  & 91.45 / 91.69   & \textbf{4.34 / 4.16}  & 0.89 / 0.88 \\
DenseGCN-14 & EfficientNet-B0 & 66.18 / 66.56  & 90.56 / 90.81   & 68.01 / 68.58   & \textbf{91.53 / 91.70}   & 1.83 / 2.02  & \textbf{0.97 / 0.89} \\ \hline
\end{tabular}}}
\label{table3}
\end{table}

\begin{table}
\centering
\caption{Ablation study on the CRC dataset, tables show the performance improvement of co-training with Deep GCN and CNN compared to training them independently.}
\renewcommand\arraystretch{1.2}{
\resizebox{\columnwidth}{!}{
\begin{tabular}{llllllll}
\hline
\multicolumn{2}{c}{\multirow{2}{*}{Modules}} & \multicolumn{6}{c}{Accuracy / Macro F1-Score of CRC}                                                       \\ \cline{3-8} 
\multicolumn{2}{c}{}                         & \multicolumn{2}{c}{Independent} & \multicolumn{2}{c}{Co-Training} & \multicolumn{2}{c}{Up $\uparrow$} \\ \hline
GCN         & CNN             & GCN           & CNN           & GCN           & CNN           & GCN         & CNN         \\ \hline
ResGCN-14   & ResNet-18       & 95.56 / 93.36 & 97.57 / 97.51 & 96.09 / 94.40 & 97.89 / 97.91 & 0.53 / 1.04 & 0.32 / 0.40 \\
ResGCN-14   & EfficientNet-B0 & 95.56 / 93.36 & 99.15 / 99.15 & \textbf{96.41 / 94.65} & \textbf{99.89 / 99.91} & \textbf{0.85 / 1.29} & \textbf{0.74 / 0.76} \\ \hline
\end{tabular}}}
\label{table4}
\end{table}

\subsection{Ablation Study}
In order to understand the role of each component of our proposed framework better, we perform ablation studies on both the LUAD7C and CRC datasets. We analyze each component individually while fixing the others, i.e., 1) asymmetric dual network architecture; 2) co-training strategy; 3) domain adaptative cell segmentation; 4) cell graph construction. It is worth noting that if none of the asymmetric dual network architecture and the co-training module are adopted, our proposed framework will degenerate into traditional single-modal GCN or CNN methods.

\subsubsection{Asymmetric Dual Network Architecture}
Our proposed asymmetric dual network architecture is efficient, low-coupling, and highly generalizable. We first compare our asymmetric co-training with independent training on the LUAD7C dataset. As is shown in \autoref{table3}, it can be observed that our multi-modal co-training architecture is superior to the traditional single-modal architecture, either the cell-based Deep GCN or the pixel-based CNN. The stable improvement of the accuracy and macro F1-score on the LUAD7C dataset shows the effectiveness of our architecture as it provides comprehensive utilization of the feature extraction capabilities of the two branch models. What‘s more, it is worth noting that we can replace our backbones in the proposed framework. Specifically, we set up four scenarios in this ablation experiment. The Deep GCN backbone can be ResGCN-14 or DenseGCN-14, and the CNN backbone can be ResNet-18 or EfficientNet-B0, which shows that our architecture is low-coupling and scalable. The results of these four different combinations fully demonstrate the robustness of our architecture in multi-class histopathological image classification tasks. Finally, we also validate our architecture on the public CRC dataset, as is shown in \autoref{table4}. The steady improvement of accuracy and macro F1-score shows good generalization ability of our proposed asymmetric dual network architecture.

\subsubsection{Co-Training Strategy}
\begin{figure}
\centerline{\includegraphics[width=\columnwidth]{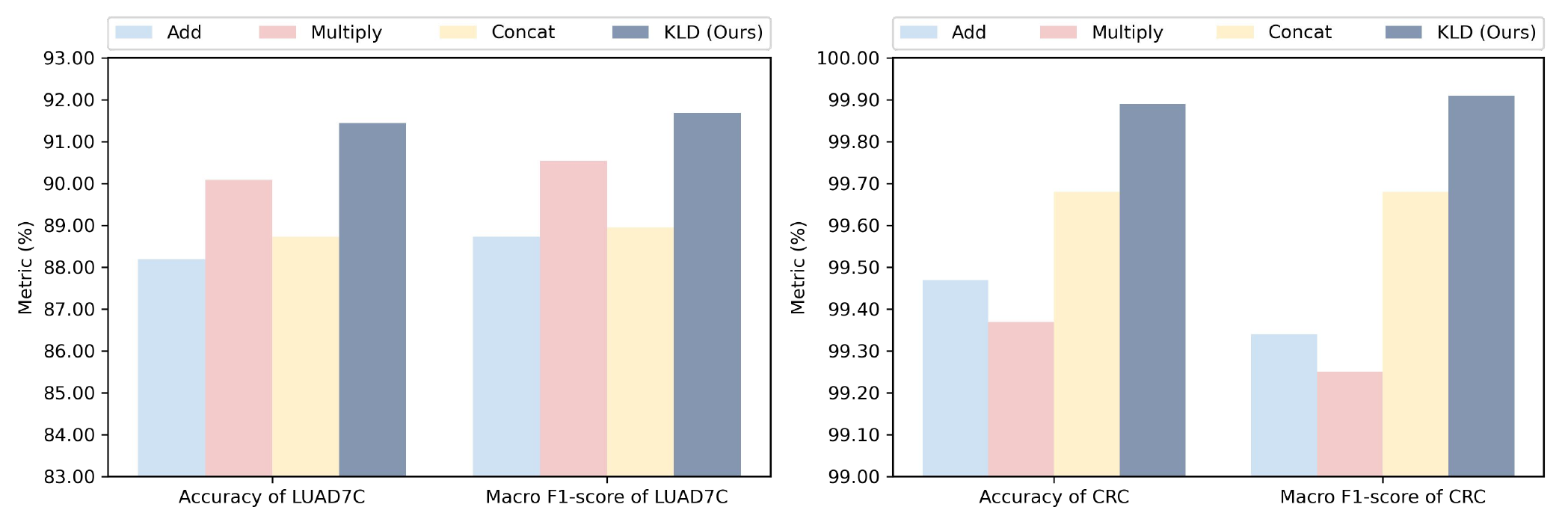}}
\caption{Ablation study of our asymmetric Co-Training strategy (KLD) on our collected LUAD7C dataset and the public CRC dataset, the figure shows the accuracy and macro F1-score comparison with other conventional model fusion methods like add, multiplication, and concatenation.}
\label{fig7}
\end{figure}

We also compare our co-training strategy, i.e., the KL divergence with threshold constraints (KLD), with traditional static model fusion methods like add, multiplication, and concatenation forms \cite{pati2022hierarchical}, shown in \autoref{fig7}. Overall, the proposed method consistently outperforms the fixed non-interactive methods since our dynamic interaction mechanism can adaptively select the useful multi-level features for the task and produce more abundant information for decision-making. By dynamically reducing the KL distance of features extracted from different modal spaces as a driving force, two different modalities can learn from each other. A bridge of complementary information is established between the pixel-level modality and the cell-graph modality. The experiment results strongly demonstrate the effectiveness of our proposed co-training strategy.

\subsubsection{Domain Adaptative Cell Segmentation}
To get more accurate cell instance segmentation, we build an unsupervised cell segmentation network using domain adaptative techniques. As is shown in \autoref{table5}, we compare our domain adaptative cell segmentation with traditional transfer learning without domain adaptation, e.g., using only U-Net. The improvement of accuracy and macro F1-score on both the LUAD7C and CRC datasets indirectly demonstrates the superiority of our domain adaptative cell segmentation. Compared to traditional cell segmentation, our domain adaptative cell segmentation effectively tackles the generalization problem between the source domain and the target domain, thus generating more accurate cell segmentation on the target domain, leading to more reliable node features in the cell graph.

\begin{table}[]
\centering
\caption{Ablation: Analysis of domain adaptative cell segmentation. Accuracy and macro F1-score on the LUAD7C and CRC datasets.}
\renewcommand\arraystretch{1.2}{
\resizebox{\columnwidth}{!}{
\begin{tabular}{lll}
\hline
\multirow{2}{*}{Cell Segmentation Module} & \multicolumn{2}{l}{Accuracy / Macro F1-score}   \\ \cline{2-3} 
                                   & LUAD7C                   & CRC                    \\ \hline
U-Net (without domain adaptation)  & 91.02 / 91.27          & 99.68 / 99.65          \\
Domain adaptative framework (Ours) & \textbf{91.45 / 91.69} & \textbf{99.89 / 99.91} \\ \hline
\end{tabular}}}
\label{table5}
\end{table}

\subsubsection{Cell Graph Construction}
To utilize both cell appearance and spatial information, we construct a non-Euclidean space data representation of cell graphs by integrating cell-level feature and position information. As is shown in \autoref{table6}, we compare our integrated non-Euclidean data representation with traditional Euclidean data representation using only the position information of cells. The results on both the LUAD7C and CRC datasets show that our cell graph construction outperformed traditional ones. Compared to Euclidean data representation, our non-Euclidean data representation incorporates more cellular organization and characteristics information, thus generating better cell graph embeddings.

\begin{table}[]
\centering
\caption{Ablation: Analysis of cell graph construction. Accuracy and macro F1-score on the LUAD7C and CRC datasets.}
\renewcommand\arraystretch{1.2}{
\resizebox{\columnwidth}{!}{
\begin{tabular}{lll}
\hline
\multirow{2}{*}{Cell Graph Construction Module} & \multicolumn{2}{l}{Accuracy / Macro F1-score}   \\ \cline{2-3} 
                                   & LUAD7C                   & CRC                    \\ \hline
Euclidean representation  & 91.13 / 91.38          & 99.79 / 99.78          \\
Non-Euclidean representation (Ours) & \textbf{91.45 / 91.69} & \textbf{99.89 / 99.91} \\ \hline
\end{tabular}}}
\label{table6}
\end{table}

\section{Conclusion}
In this paper, we propose an asymmetric co-training framework that fuses an explainable Deep GCN and a high-performance CNN to address the challenging yet important multi-class histopathological image classification. The proposed method comprehensively utilizes the pixel-level features and cell-level features in dynamic interactions, thus outperforming single-modal methods. The auxiliary co-training of Deep GCN makes the entire framework more explainable by emphasizing the notion of biological tissue entities, their topological distribution, and inter-entity interactions. The private clinically acquired dataset LUAD7C, including seven subtypes of lung adenocarcinoma, is rare and much more challenging in multi-class histopathological image classification tasks. Our proposed method achieves state-of-the-art on the private LUAD7C dataset as well as the public CRC dataset. Experimental results demonstrate the effectiveness, explainability, and generalization of our proposed framework in multi-class histopathological image classification.

\bibliographystyle{ieeetr}
\bibliography{tmi}

\end{document}